\documentclass[letterpaper, 10 pt, conference]{ieeeconf}  % Comment this line out if you need a4paper

\IEEEoverridecommandlockouts                              % This command is only needed if 
\usepackage{graphics} % for pdf, bitmapped graphics files
\usepackage{wrapfig}
\usepackage{kotex}
\usepackage{graphicx}
\usepackage{tabularx}
\usepackage{subcaption}
\usepackage{cite}
\usepackage{romannum}
\usepackage{amsmath,amssymb}
\usepackage{url}
\usepackage{bm}
\usepackage[table]{xcolor}

\usepackage[ruled,vlined]{algorithm2e}
\usepackage[noend]{algpseudocode}
\makeatletter
\let\OldStatex\Statex
\renewcommand{\Statex}[1][3]{%
  \setlength\@tempdima{\algorithmicindent}%
  \OldStatex\hskip\dimexpr#1\@tempdima\relax}
\renewcommand{\ALG@beginalgorithmic}{\normalsize}

\newcommand{\vb}{{\bf b}}

\newcommand{\vx}{{\bf x}}

\newcommand{\vz}{{\bf z}}
\newcommand{\vo}{{\bf o}}

\newcommand{\vu}{{\bf u}}

\newcommand{\vF}{{\bf F}}

\newcommand{\vW}{{\bf W}}

\makeatother
\DeclareCaptionFont{mysize}{\fontsize{8}{9.6}\selectfont}
\title{\LARGE \bf
Physics Embedded Neural Network Vehicle Model and Applications in Risk-Aware Autonomous Driving Using Latent Features}

\author{Taekyung Kim$^*$, Hojin Lee$^*$, and Wonsuk Lee$^\dagger$% <-this % stops a space
\thanks{The authors are with the Ground Technology Research Institute, Agency for Defense Development, Daejeon 34186, Republic of Korea. \texttt{\{ktk1501, hojini1117, wsblues\}@add.re.kr}}%
\thanks{$^{*}$These authors contributed equally to this work}%
\thanks{$^{\dagger}$Corresponding author}%
}

\begin{document}
\maketitle
\thispagestyle{empty}
\pagestyle{empty}

%%%%%%%%%%%%%%%%%%%%%%%%%%%%%%%%%%%%%%%%%%%%%%%%%%%%%%%%%%%%%%%%%%%%%%%%%%%%%%%%
\begin{abstract}
Non-holonomic vehicle motion has been studied extensively using physics-based models. Common approaches when using these models interpret the wheel/ground interactions using a linear tire model and thus may not fully capture the nonlinear and complex dynamics under various environments. On the other hand, neural network models have been widely employed in this domain, demonstrating powerful function approximation capabilities. However, these black-box learning strategies completely abandon the existing knowledge of well-known physics. In this paper, we seamlessly combine deep learning with a fully differentiable physics model to endow the neural network with available prior knowledge. The proposed model shows better generalization performance than the vanilla neural network model by a large margin. We also show that the latent features of our model can accurately represent lateral tire forces without the need for any additional training. Lastly, We develop a risk-aware model predictive controller using proprioceptive information derived from the latent features. We validate our idea in two autonomous driving tasks under unknown friction, outperforming the baseline control framework.
\end{abstract}

%%%%%%%%%%%%%%%%%%%%%%%%%%%%%%%%%%%%%%%%%%%%%%%%%%%%%%%%%%%%%%%%%%%%%%%%%%%%%%%%
\section{Introduction}

In recent years, Model Predictive Control (MPC) has shown promising results for autonomous driving tasks under complex and nonlinear environments \cite{williams_information_2017, kabzan_learning-based_2019}. Typically, MPC schemes optimize an open-loop control sequence and then execute it until the next optimization update. With such a strategy, acquiring accurate models of dynamical systems is critical. Inaccurate long-horizon predictions from the model will result in an incorrectly optimized trajectory and eventually lead to inappropriate motions.

Extensive studies have sought to create analytic models of non-holonomic vehicle motion, and such works can be divided into two broad categories: kinematic models and dynamic models. Kinematic models that assume no wheel slip are insufficient to represent the motions of a high-speed driving vehicle \cite{rajamani_vehicle_2011}. Despite attempts by dynamics models to describe the complex wheel/ground interactions by utilizing a linear tire model, they reveal their natural limitations when the environments are highly nonlinear \cite{kong_kinematic_2015}.

To this end, recent robotics studies have been widely employing machine learning models owing to their universal function approximation properties. Most standard machine learning techniques to learn dynamics models include a Gaussian Process and a deep-learning-based approach using neural networks \cite{nguyen-tuong_model_2011}. In particular, current neural networks use different architectures to approximate the model, e.g., deterministic models \cite{williams_information_2017, nagabandi_neural_2018}, stochastic models \cite{lambert_low-level_2019}, and ensemble models \cite{janner_when_2019}. However, these black-box learning approaches completely ignore the existing knowledge of well-known physics and create the mapping between inputs and states from scratch. This strategy not only makes the learning problem unnecessarily more difficult but also cannot ensure physical plausibility. 

In this paper, we seamlessly combine deep learning with a fully differentiable physical vehicle model to endow the neural network with available prior knowledge. We also present a control scheme that can be applied in conjunction with our neural network model to perform risk-aware autonomous driving tasks with unknown friction.

\begin{figure*}[t]
\centering
\includegraphics[width=0.9\textwidth]{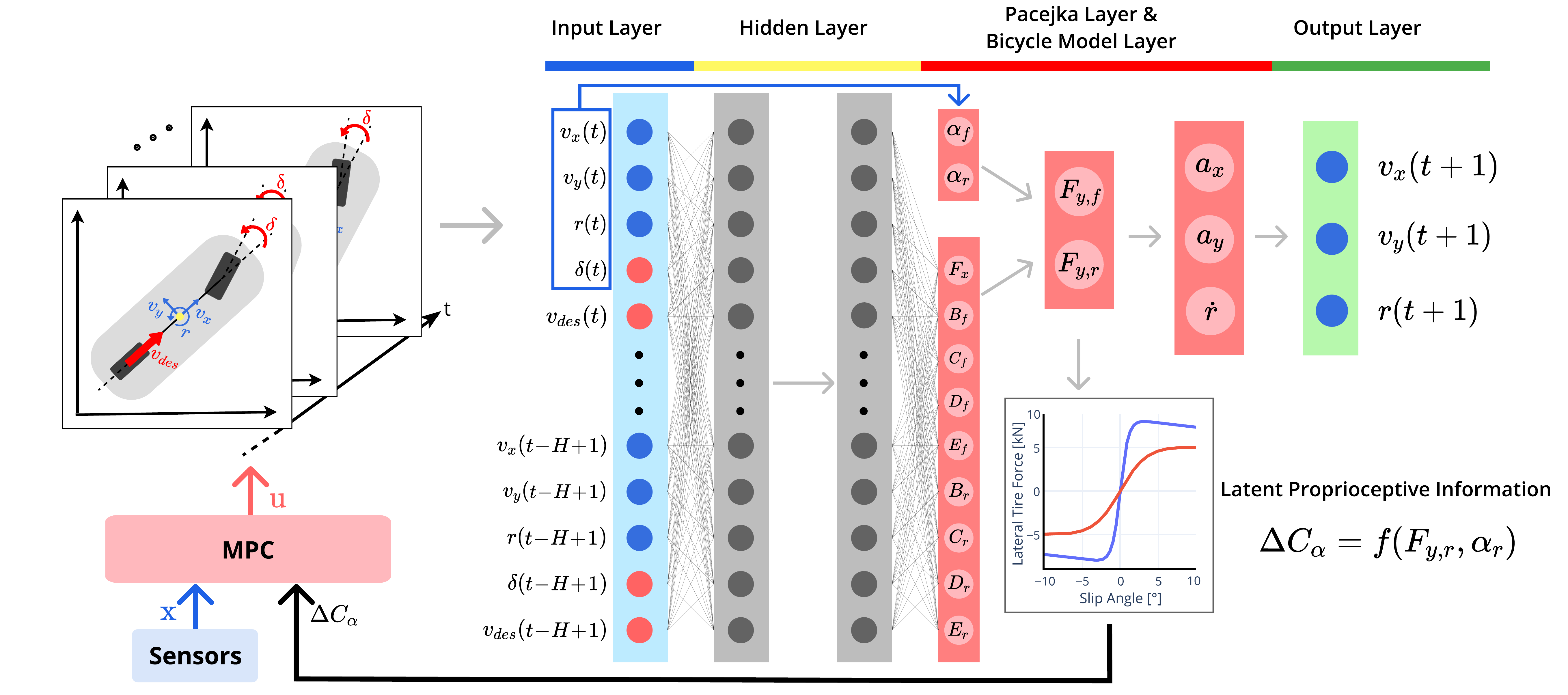}
\caption{Overview diagram of our risk-aware autonomous driving strategy using the latent proprioceptive information obtained from the physics-embedded neural network dynamics.}
\label{fig:Pacejka_flow_chart}
\vspace*{-0.05in}
\end{figure*}

\subsection{Related Work}

While model learning for dynamic systems has been extensively studied, only a few approaches use physics knowledge to improve learning performance outcomes. One study \cite{nguyen-tuong_using_2010} proposed the use of knowledge of the physical model to become part of the mean function or of the kernel in a Gaussian Process setting. In this approach, however, the representation of the model still remains a black-box.

A more recent development is the Physics-Informed Neural Network (PINN) \cite{raissi_physics-informed_2019}, which structures the learning problem by involving the knowledge of a specific differential equation, leading to lower sample complexity. This method can directly solve a specific set of differential equations after training. 

One line of research shares the motivation of the PINN but focuses instead on model learning problems \cite{lutter_deep_2019, greydanus_hamiltonian_2019, cranmer_lagrangian_2020}. These methods approach the model learning problem using a structure that infers equations of motion. These approaches do not require to be engineered according to physical assumptions specific to each system embodiment and are generally applicable to a wide range of mechanical systems. On the other hand, we focus on the fact that if there are existing physical priors extensively addressed by prior work, explicitly embedding them into the neural network could yield extra benefits.
\subsection{Contributions}

In this paper, we propose to embed a well-known physics-based model into a neural network to capture the underlying vehicle dynamics. We add the Pacejka magic formula \cite{pacejka_magic_1992} and the nominal dynamics of the bicycle model as the last hidden layers. In the middle of forward propagation of this model, we can extract latent variables representing the lateral tire forces that are difficult to measure in the real world. We leverage these latent features as hints to be given for the risk-aware MPC. To the best of the authors' knowledge, there has never been a control scheme in which an MPC adapts to the latent variables of a physics-inspired neural network. An overall diagram of the proposed control scheme is illustrated in Fig.~\ref{fig:Pacejka_flow_chart}.

In summary, the advantages of using the proposed framework are as follows:
\begin{itemize}
    \item The model learning problem is entailed by physics priors to yield a faster learning speed, higher accuracy, and better generalization.
    \item The proposed model can accurately estimate lateral tire forces with zero effort without the need for any additional training with ground truth data.
    \item The latent proprioceptive information, which can be acquired from the forward propagation of the model during the MPC optimization process, can be used as valuable hints when designing a risk-aware autonomous driving controller with unknown friction. \footnote{Our video can be found at: \url{https://youtu.be/rPN0hGCW0R4}}
\end{itemize}

\section{Background}
\begin{figure}[t]
\centering
\includegraphics[width=0.85\linewidth]{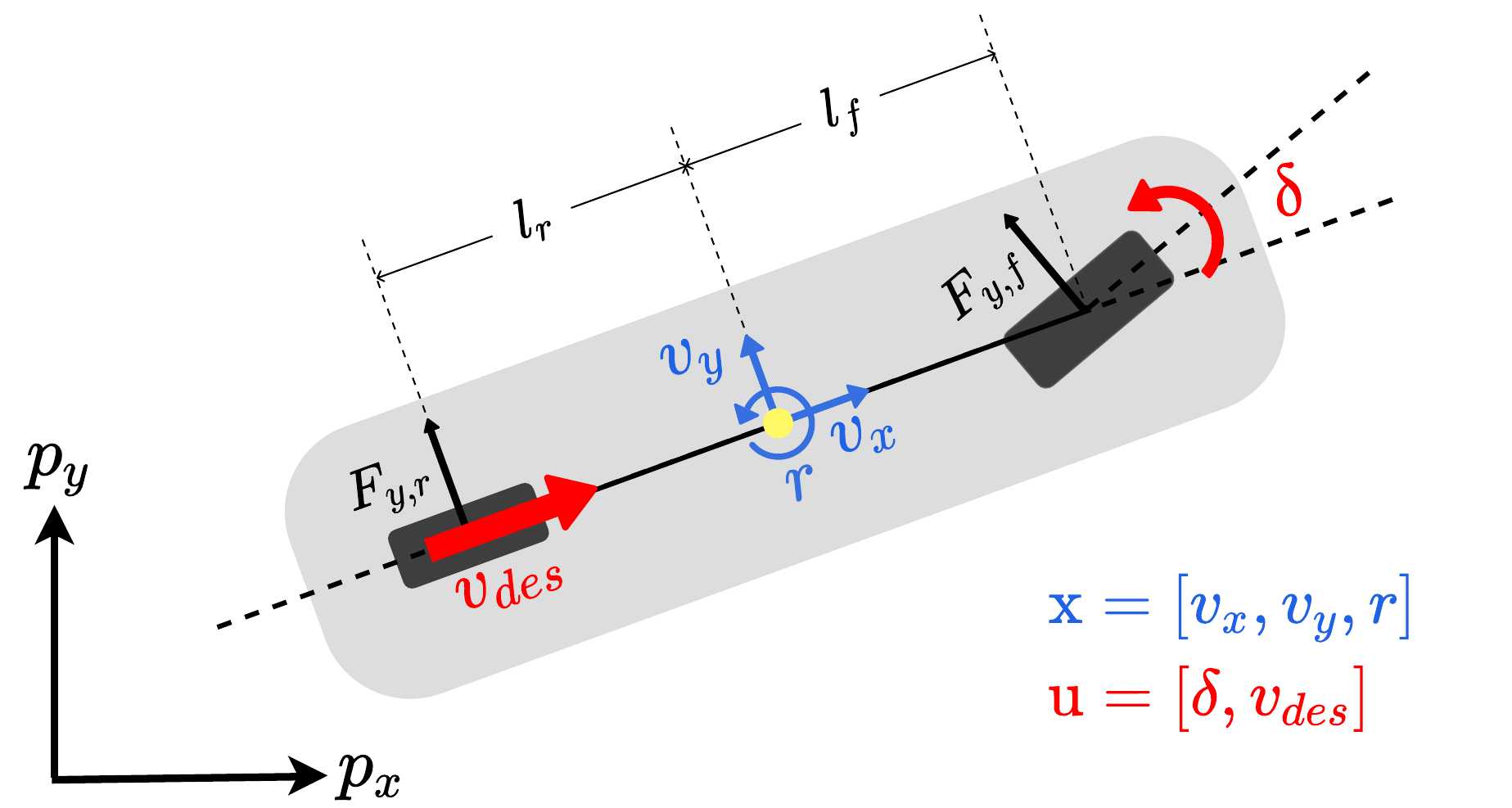}
\caption{A diagram of the bicycle model. The dynamic state variable $\vx$ is depicted in blue. The input variable $\vu$ is depicted in red.}
\label{fig:bicycle_model}
%\vspace*{-0.05in}
\end{figure}

In the following, we provide a brief overview of the Pacejka tire model \cite{pacejka_magic_1992}. To model the motions of a vehicle driving at high speeds, the dynamic bicycle model is commonly employed because it provides a good trade-off between model accuracy and simplicity for real-time implementation \cite{kabzan_learning-based_2019}. The bicycle model is illustrated in Fig.~\ref{fig:bicycle_model}. The dynamic state variable $\vx = \left(v_x, v_y, r\right)^{\top}$ consists of the longitudinal velocity $v_x$, the lateral velocity $v_y$, and the yaw rate $r$. The kinematic state variable in the global coordinate frame consists of the position $\left(p_x, p_y\right)^{\top}$ and the heading angle $\theta$. The inputs to the system $\vu = \left(\delta, v_{des}\right)^{\top}$ are the steering angle $\delta$ and the desired speed $v_{des}$. Note that, unlike the electric vehicles used in recent MPC-based autonomous driving studies \cite{williams_aggressive_2016, kabzan_learning-based_2019}, the majority of current vehicles shift gears automatically. The throttle is undesirable to be used as an action in these vehicles because the dynamic characteristics change dramatically as the gears shift. Therefore, we here use $v_{des}$ as the input and append a low-level Proportional-Integral (PI) velocity controller to manage the throttle and brake.

The distances between the center of gravity and the front and rear tires are denoted as $l_f$ and $l_r$, respectively. The tire sideslip angles can be obtained as follows:
\begin{equation}
\begin{aligned}
    \alpha_f & = \arctan \left( \frac{v_y + l_f{r}}{v_x}\right) - \delta, \\
    \alpha_r & = \arctan \left( \frac{v_y - l_r{r}}{v_x}\right),
\end{aligned}
\label{eq:slip}
\end{equation}
where the notations $(\cdot)_f$ and $(\cdot)_r$ denote correspondingly the front and the rear of the vehicle. Then, the dynamics of the tire-road contacts can be formulated into nonlinear equations:
\begin{equation}
\begin{aligned}
    F_{y,f}  = F_z{D_f}&\sin({C_f}\arctan(B_f{\alpha_f}-\\
    & E_f(B_f{\alpha_f}- \arctan(B_f{\alpha_f})))), \\
    F_{y,r}  = F_z{D_r}&\sin({C_r}\arctan(B_r{\alpha_r}- \\
    & E_r(B_r{\alpha_r}-\arctan(B_r{\alpha_r})))),
\end{aligned}
\label{eq:pacejka}
\end{equation}
where $F_z$ denotes the vertical load of the tire. The model parameter $B$ refers to the stiffness factor, $C$ refers to the shape factor, $D$ denotes the peak value, and $E$ is the curvature factor \cite{rajamani_vehicle_2011}. 

Conventional engineering approaches would determine these parameters through a system identification approach with real-world driving tests. Unfortunately, the test setup for these measurements is complicated and costly, and the data-collection procedures by manual driving are time-consuming tasks. Furthermore, they are determined under a specific set of conditions and naturally change as environmental factors such as road friction change. These factors complicate the use of this tire model in an autonomous driving task under unknown friction. This problem is also related to the fact that the lateral tire forces $F_{y}$ are difficult to be obtained analytically.

\section{Physics Embedded Neural Network Vehicle Model}
\subsection{The Baseline Model}
Our research goal is to control an Unmanned Ground Vehicle (UGV) so that it drives successfully under unknown surfaces with high and low friction. Accurate dynamics modeling must be accompanied to achieve such control, but the complex nonlinear motions in this environment are difficult to model. Depending on the friction level of the road surface, the same lateral force applied to the wheel at the same corner will result in completely different vehicle motions. There is one more factor that adds complexity to the modeling problem, which is the time-varying characteristics due to the automatic gear shifting. 

To overcome these issues, we use a neural network in which historical state-action pairs $\bm{\mathcal{H}}$ are used as inputs to capture time-varying behaviors \cite{spielberg_neural_2019}:
\begin{equation}
    \bm{\mathcal{H}}_t = \left[ \vx_t, \,  \vu_t, \, \vx_{t-1}, \,  \vu_{t-1}, \, \dots \vx_{t-H+1}, \,  \vu_{t-H+1} \right] .
\label{Equation:history_dynamics}
\end{equation}
The length of the history is denoted as $H$ and the discretized time variable is denoted as $t$. This strategy enables the network to accurately approximate nonlinear and complex dynamics without explicitly encoding the automatic gear shifting mechanism \cite{kim_smooth_2022}. Then, we can employ a standard fully-connected neural network $\vF$ to predict the discrete change in states over the time step duration of $\Delta t$ \cite{lambert_low-level_2019}:
\begin{equation} 
    \vx_{t+1}-\vx_{t} = \vF(\bm{\mathcal{H}}_t).
\end{equation}
The idea behind estimating the residual difference in states is proposed to be effective when using a small discretized time interval \cite{nagabandi_neural_2018}.

Following prior work \cite{kim_toast_2022}, we design a baseline feedforward neural network (abbreviated as ``FF-NN") consisting of four fully-connected hidden layers with $H=4$. The input to each activation is denoted using $\vz_\diamond$ and the output of the activation is indicated with $\vo_\diamond$, where $\diamond$ represents the layer number. The activation function is denoted using $\operatorname{act}(\vz_\diamond)$. $\vW_\diamond$ and $\vb_\diamond$ are correspondingly the weights and biases of the network. The baseline model then takes the following form:
\begin{align}
\vz_{1} &=\vW_{1}^{\top} \bm{\mathcal{H}}_{t}+\vb_{1} \\
\vo_{1} &=\operatorname{act}\left(\vz_{1}\right) \\
& \qquad \dots \nonumber\\
\vo_{3} &=\operatorname{act}\left(\vz_{3}\right) \\
\vz_{4} &=\vW_{4}^{\top} \vo_{3}+\vb_{4} \label{eq:last_hidden} \\
\vx_{t+1} &= \vx_t + \vz_{4} .
\end{align}
\subsection{Deep Pacejka Model}

In this section, we introduce our \textit{Deep Pacejka Model} (DPM). We propose to explicitly embed the prior knowledge of the vehicle physics model into the neural network. We add the Pacejka tire model as a layer between the last hidden layer of the baseline network and its output layer. Here, each node in the last hidden layer (\ref{eq:last_hidden}) predicts the parameters of the physics model in (\ref{eq:pacejka}). Note that the longitudinal force $F_x$ applied to the vehicle is an essential component of the vehicle's lateral dynamics. However, unlike an electric vehicle, it is undesirable to model it as a linear relationship with throttle commands in a vehicle with an automatic transmission. Therefore, we also let the neural network predict this difficult-to-model force:
\begin{equation}
{\left[F_x, B_f, C_f, D_f, E_f, B_r, C_r, D_r, E_r \right]}^{\top}= \vz_{4} .
\end{equation}

Then, the physics model (\ref{eq:pacejka}) is followed by the last hidden layer to compute the lateral forces of both the front and rear tires. We denote this layer as the \textit{Pacejka layer}. Subsequently, the output of the Pacejka magic formula is forward propagated with the longitudinal force into the nominal dynamics of the bicycle model \cite{kong_kinematic_2015}:
\begin{equation}
\arraycolsep=1.4pt\def\arraystretch{1.5}
\dot{\vx}_{t+1}=\left[\begin{array}{c}
\frac{1}{m}\left(F_x - F_{y,f}\,{\sin(\delta(t))} + m{v_y(t)}{r(t)}\right) \\
\frac{1}{m}\left(F_{y,r} + F_{y,f}\,{\cos(\delta(t))} - m{v_x(t)}{r(t)}\right) \\
\frac{1}{I_z}\left(F_{y,f}\,{\cos(\delta(t))}\,{l_f} - F_{y,r}\,{l_r}\right)
\end{array}\right] ,
\label{eq:bicycle}
\end{equation}
where $m$ and $I_z$ denote the vehicle's mass and yaw moment of inertia, respectively, and they are determined by the vehicle. We denote this layer as the \textit{bicycle model layer}. Finally, we can obtain the next state using the state derivative:
\begin{equation}
\vx_{t+1} = \vx_{t} + \dot{\vx}_{t+1} \, {\Delta {t}}.
\end{equation}

Here, we emphasize that the DPM model still outputs the residual difference in states, identical to the baseline network. In addition, the Pacejka formula (\ref{eq:pacejka}) and the nominal dynamics (\ref{eq:bicycle}) are fully differentiable. This allows the DPM to be trained in an end-to-end manner by minimizing precisely the same objective used in the baseline network, such that $\left|\left|\vx_{t+1}-\vx_{t}-\vF(\bm{\mathcal{H}}_t)\right|\right|^2$. Lastly, our method is a general solution that can be utilized in other types of model architectures, including Recurrent Neural Networks (RNN). Furthermore, it can be formulated as a stochastic model or an ensemble model. An ablation study of these considerations is, however, out of the scope of this paper.

In one line of research, the model learning problem is structured generally with physical priors, but it does not require knowledge of the specific physical embodiment \cite{lutter_deep_2019, greydanus_hamiltonian_2019, cranmer_lagrangian_2020}. Conversely, we explicitly embed system-specific physics into the neural network as layers, inspired by certain ideas introduced in vehicle trajectory prediction studies \cite{tang_adaptive_2019, cui_deep_2020}. The advantage of taking this difference is that the forward propagation of the model can produce latent variables that are typically difficult to obtain in the real world. In our setting, we can implicitly acquire the lateral tire forces as the output of the \textit{Pacejka layer}. Note that directly measuring these forces requires wheel force transducers, which are very expensive sensors. An alternative way is to design a dedicated observer to estimate them \cite{baffet_observer_2007, acosta_tire_2018}. In contrast, the proposed method can obtain them with zero effort.

\subsection{Estimation Results \label{section:nn_result}}

Throughout this paper, we use IPG Carmaker, a high-fidelity vehicle simulator that solves nonlinear dynamics in real time, to validate our idea. We collected $70$ minutes of training data by manually driving under various conditions, with a data rate of $10$ Hz ($\Delta t = 0.1$ s). The dataset consists of four different types of maneuvers: 1) zig-zag driving at low speeds, 2) high-speed driving, 3) driving to slide as much as possible, and 4) random motions. 1) and 2) were collected on a race track with six sharp corners. 3) and 4) were collected on flat ground. The maximum vehicle speed $v_{max}$ was set to $40$ km/h. These maneuvers were performed on seven different road surfaces, each with a distinct friction coefficient: $\mu \in \{0.4, 0.5, \dots 1.0\}$. A Volvo XC90 was used as the control vehicle.

We trained the FF-NN and the DPM for comparison with a fixed random seed. Except for the physics embedding, the same settings were applied to both models. We split the data into the training and test sets with a 7:3 ratio. A Rectified Linear Unit (ReLU) was used for activation. The Mean Squared Error (MSE) loss was used as the loss function and the networks were trained with the Adam optimizer. We applied an early stop on the test set. 
\begin{wrapfigure}{R}{0.5\linewidth}
\centering
\includegraphics[width=0.99\linewidth]{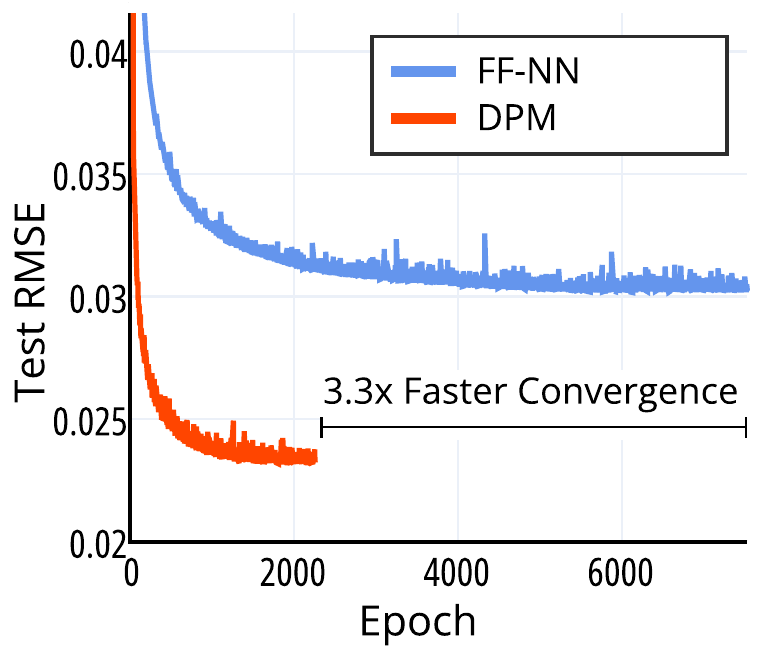}
\caption{The test RMSE curves for the compared methods during training.}
\label{fig:test_rmse}
\end{wrapfigure}

We use two error metrics for the performance evaluation: the root mean square error (denoted as RMSE) and the max error (denoted as $E_{max}$). The resulting test error curves for both methods are shown in Fig.~\ref{fig:test_rmse}. The results show that the proposed method converged much faster than the FF-NN because the physics prior had guided the learning problem. Our method also converged to a better solution than the FF-NN. Concretely, the test errors for each state element are shown in Table~\ref{table:test_errors}.

\begin{table}[ht]
\renewcommand\arraystretch{1.2}
\captionsetup{justification=centering}
\caption{}
\caption*{\textsc{Estimation results using test dataset after training.}}
\label{table:test_errors}
\begin{center}
\begin{tabularx}{1.0\columnwidth}{c|cc|cc|cc|}
& \multicolumn{2}{c|}{$v_x$ [m/s]} & \multicolumn{2}{c|}{$v_y$ [m/s]} & \multicolumn{2}{c|}{$r$ [rad/s]} \\ \cline{2-7}
& RMSE & $E_{max}$ & RMSE & $E_{max}$ & RMSE & $E_{max}$  \\ \cline{1-7}
FF-NN & \textbf{0.0293} & 0.4451 & 0.0355 & 0.6793 & 0.0289 & 0.5956 \\
DPM & 0.0299 & \textbf{0.3765} & \textbf{0.0214} & \textbf{0.4889} & \textbf{0.0169} & \textbf{0.4727} \\
\end{tabularx}
\end{center}
\vspace*{-0.15in}
\end{table}

We also collected validation data to demonstrate the superiority of our method in terms of generalization performance. The data were collected on the same race track used to obtain the training data, but with adjusted friction levels. The following values were assigned to the friction coefficients of each corner, with none included in the training data: $[0.45, 0.55, 0.65, 0.75, 0.85, 0.95]$. The other regions were set to $\mu = 1.0$. Table~\ref{table:val_errors} shows the validation errors for each state element in detail. It's worth noting that lateral dynamics estimations ($v_y$ and $r$) are more critical for safe vehicle control; this demonstrates that the DPM has greater generalization capabilities than the FF-NN by a large margin.

\begin{table}[ht]
\renewcommand\arraystretch{1.2}
\captionsetup{justification=centering}
\caption{}
\caption*{\textsc{Estimation results using validation dataset after training.}}
\label{table:val_errors}
\begin{center}
\begin{tabularx}{1.0\columnwidth}{c|cc|cc|cc|}
& \multicolumn{2}{c|}{$v_x$ [m/s]} & \multicolumn{2}{c|}{$v_y$ [m/s]} & \multicolumn{2}{c|}{$r$ [rad/s]} \\ \cline{2-7}
& RMSE & $E_{max}$ & RMSE & $E_{max}$ & RMSE & $E_{max}$  \\ \cline{1-7}
FF-NN & \textbf{0.0224} & \textbf{0.3136} & 0.0265 & 0.2358 & 0.0225 & 0.2179 \\
DPM & 0.0234 & 0.3281 & \textbf{0.0128} & \textbf{0.1106} & \textbf{0.0122} & \textbf{0.1220} \\
\end{tabularx}
\end{center}
\vspace*{-0.15in}
\end{table}

\begin{figure}[t]
\centering
\includegraphics[width=0.95\linewidth]{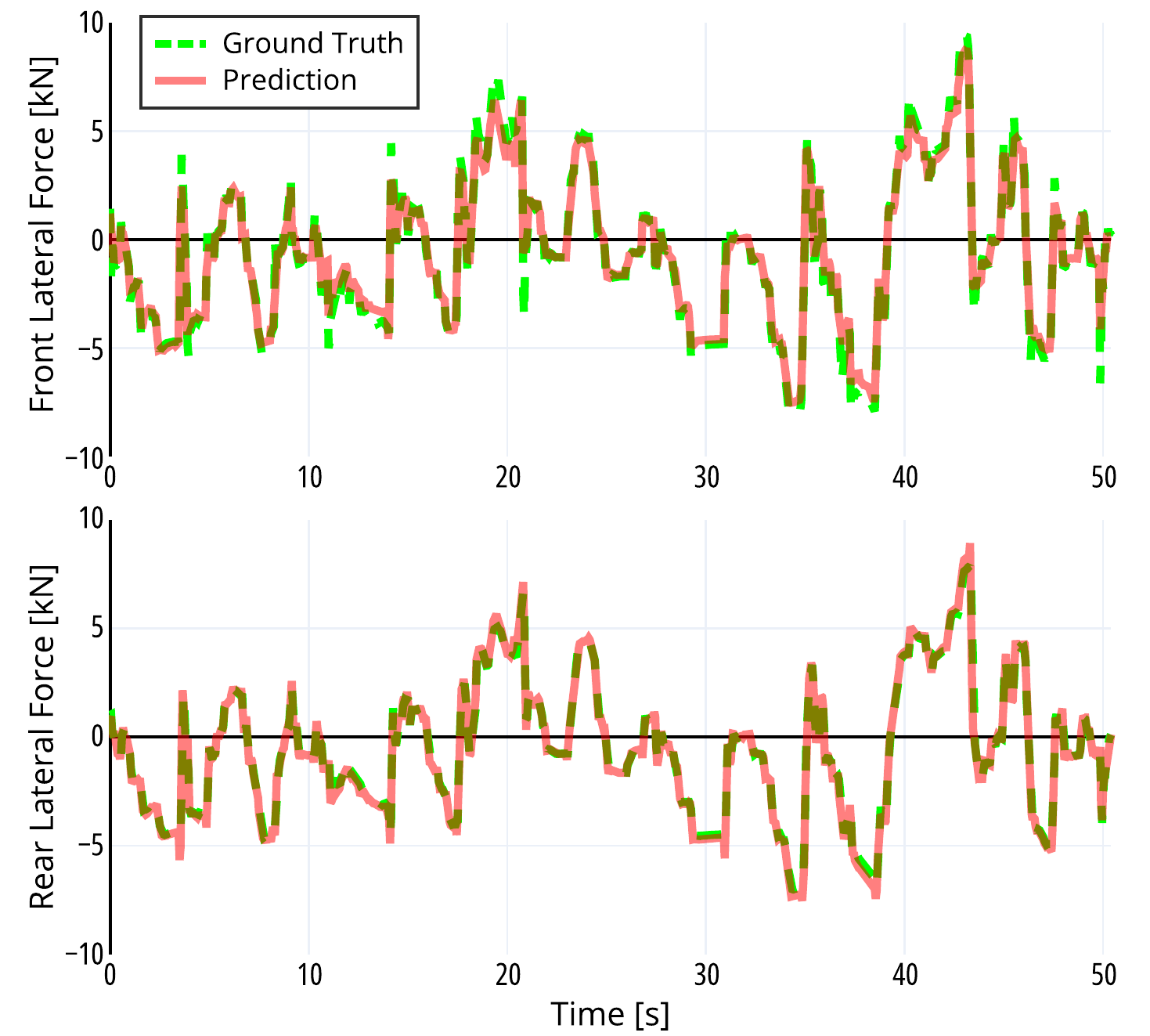}
\caption{The prediction results of lateral tire forces on the validation data.}
\label{fig:fy_prediction}
\end{figure}

\begin{figure*}[t]
\centering
\includegraphics[width=0.98\textwidth]{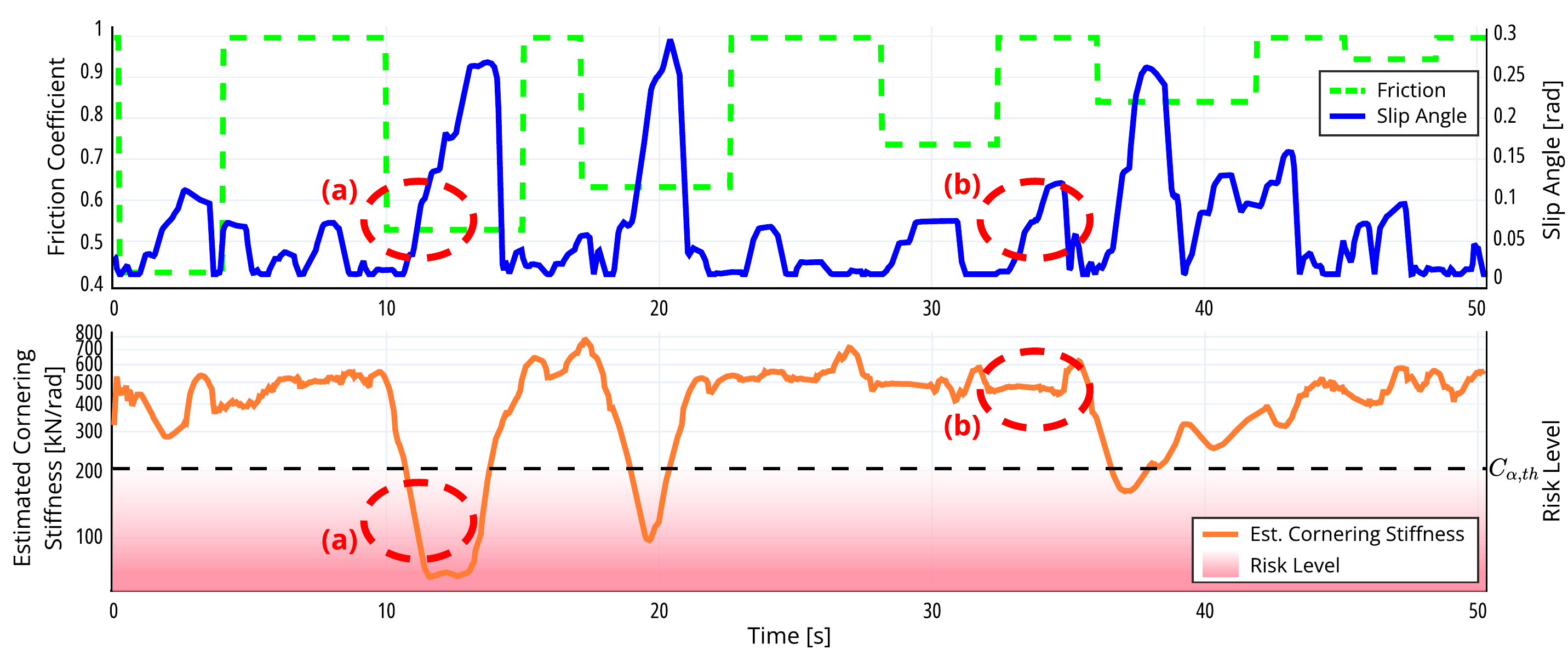}
\caption{The estimation results of the cornering stiffness on the validation data (in orange). The ground truth friction coefficient of the road surface (in green) and the vehicle slip angle (in blue) are also visualized. Note that we apply logarithmic scaling to the cornering stiffness.}
\label{fig:estimated_bcd}
% \vspace*{-0.1in}
\end{figure*}

Lastly, we visualize the latent features of lateral tire forces $F_y$ while the DPM estimates with the validation data (see Fig.~\ref{fig:fy_prediction}). The results show that the extracted latent features closely match the ground truth values from the simulator. It is important to point out that the model was never trained with respect to $F_y$. Despite the absence of ground truth data for supervised learning, our method can successfully estimate the parameters of the physics model, resulting in accurate predictions on difficult-to-model forces.

\subsection{Extracting Latent Information for Risk-Aware Control \label{section:latent}}

\begin{figure}[t]
\centering
\includegraphics[width=0.95\linewidth]{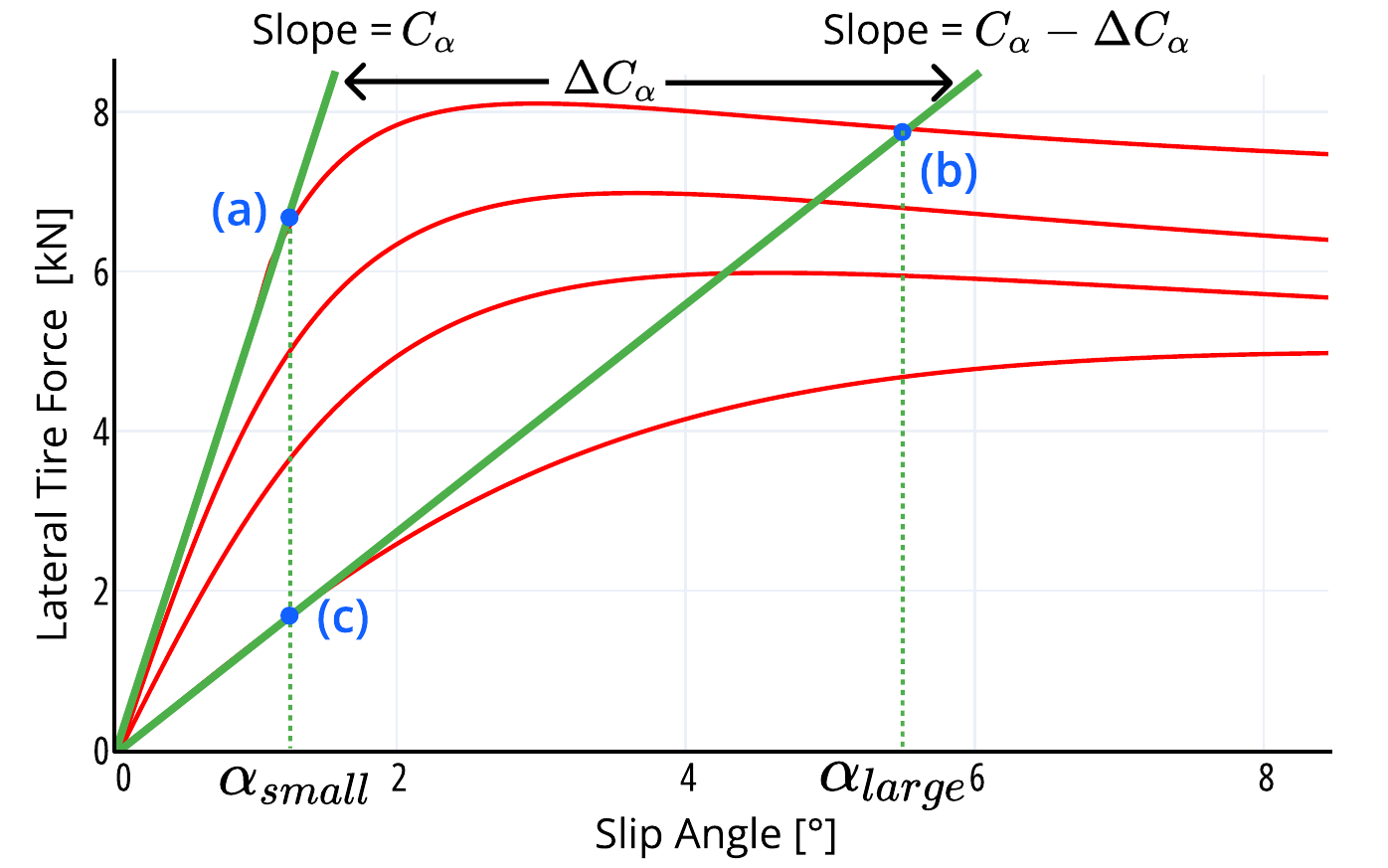}
\caption{Lateral tire force models with different environments.}
\label{fig:bcd_curve}
\end{figure}

In this section, we describe the method of leveraging the latent variables to provide useful information to the controller. The relationship curve between the lateral tire force and sideslip angle under different friction levels is shown in Fig.~\ref{fig:bcd_curve}. As long as the tire slip angle is small, the lateral tire force equation can be approximated by linearizing (\ref{eq:pacejka}) as $F_y \approx C_{\alpha} \, \alpha$, where $C_\alpha$ is the product of the tire parameters $B \cdot C \cdot D$ and generally referred to as the cornering stiffness. Although this approximation is effective for normal driving tasks in urban environments, it is obvious that when the tire sideslip angle is large, the lateral force begins to slip into the nonlinear area. 

We aim at being sensible of situations in which the controlled vehicle may lose its stability if the controller outputs more aggressive commands at the critical moment. Here, we assume the following two cases: 1) situations in which the lateral tire force is saturated into nonlinear regions despite a high friction level (see Fig.~\ref{fig:bcd_curve}b), and 2) situations in which the cornering stiffness decreases due to a low friction level, regardless of the slip angle (see Fig.~\ref{fig:bcd_curve}c). Both of these cases can be characterized by a decrease in $F_{y,r} / \alpha_{r}$. It can be seen in (\ref{eq:slip}) that the equation of the front slip angle contains the steering command, which can cause the resulting value to be noisy. Therefore, we use the rear axle to define this information.

The estimated cornering stiffness $F_{y,r} / (\alpha_{r} + \epsilon)$ while the DPM estimates with the validation data can be observed in Fig.~\ref{fig:estimated_bcd} ($\epsilon \to 0$). The result shows that the estimated cornering stiffness drops at corners with low friction coefficients. We propose to use this variable as a latent risk metric. It is important to point out that the slip angle is not an alternative to the proposed metric. The proposed risk metric reacts differently depending on the road surface even at the same slip angle and therefore it can quickly gain foresight into risky situations even with a small slip angle (See Fig.~\ref{fig:estimated_bcd}a and Fig.~\ref{fig:estimated_bcd}b).

Taking cornering stiffness variations into account, the linearized tire model can be expressed as follows:
\begin{equation}
F_{y,r} = \left(C_{\alpha, th} - \Delta C_{\alpha}\right) \alpha_{r} ,
\end{equation}
where $\Delta C_{\alpha}$ is the uncertainty variables of the cornering stiffness \cite{baffet_estimation_2008}, and $C_{\alpha, th}$ is a threshold variable determined empirically. Finally, we define the latent proprioceptive information yields by the DPM as shown below:
\begin{equation}
\Delta C_{\alpha} = 
    \begin{cases}
    0 , & \mbox{if }C_{\alpha, th} <  \left(\frac{F_{y,r}}{\alpha_{r} + \epsilon}\right) \\
    C_{\alpha, th} - \left(\frac{F_{y,r}}{\alpha_{r} + \epsilon}\right), & \mbox{otherwise} .
    \end{cases}
\end{equation}

\section{Risk-Aware Controller Using Sampling-Based MPC}
\subsection{Smooth Model Predictive Path Integral}

In this section, we provide a brief overview of the Smooth Model Predictive Path Integral (SMPPI) algorithm \cite{kim_smooth_2022}. Sampling-based MPC has been proposed to optimize both convex and non-convex objectives using the forward dynamics $\vF$. In particular, the Model Predictive Path Integral (MPPI) algorithm introduced a powerful technique for evaluating thousands of sample trajectories in a parallel manner using Graphic Processing Units (GPUs) \cite{williams_information_2017}. Then, an optimal control sequence is produced by calculating the weighted average of the sampled trajectories, with the weights defined by the cost of each trajectory rollout. Due to its appealing characteristics, MPPI has been successfully applied in various robotic applications.

However, the stochastic nature of such a sampling-based algorithm inevitably causes chattering in the resulting commands. This problem becomes more prominent when the MPPI attempts to respond to a rapidly changing environment. SMPPI is an extended study of the MPPI algorithm, seamlessly combined with an input-lifting strategy. This framework not only mitigates undesirable chattering but can also effectively adapt to situations in which the optimal control distribution deviates significantly from the previous iteration \cite{kim_smooth_2022}.

Following our prior work \cite{kim_smooth_2022}, the state-dependent cost function is defined as follows:
\begin{equation} \label{eq:cost}
    c(\vx) = {\text{Track}(\vx)} + {\text{Slip}(\vx)} + {\text{Speed}(\vx)} .
\end{equation}
The track cost guides the vehicle to drive inside the track boundary by imposing an impulse-like penalty. The given two-dimensional cost map $\textbf{M}$ returns 1 if the vehicle position is on the outside boundary of the track, and otherwise returns 0.
\begin{equation}
    \text{Track}(\vx) = {10000\textbf{M}(p_x,p_y)} .
\end{equation}
The slip cost imposes both soft and hard costs on the slip angle to encourage the controller to plan a stable future trajectory. $\alpha_{th}$ is an empirically determined threshold.
\begin{align}
    \text{Slip}(\vx) &= \alpha^2 + 10000I \, (\{\left | \alpha \right | > \alpha_{th}\}) \\
    \alpha &= -\text{arctan} \left ( \frac{v_y}{\lVert v_x \rVert} \right ) , \nonumber
\end{align}
where $I$ is an indicator function. The speed cost is a standard quadratic function to achieve the reference vehicle speed:
\begin{equation}
    \text{Speed}(\vx) = (v_x - v_{ref})^2 .
\end{equation}

\subsection{Velocity-Dependent Control}

To design a general risk-aware controller for autonomous driving tasks under unknown friction, we leverage the latent proprioceptive information $\Delta C_{\alpha}$ introduced in Section~\ref{section:latent}. We found that the SMPPI algorithm is an effective baseline for this purpose. The sampling-based MPC offers significant advantages in terms of objective function flexibility, allowing for easy encoding of high-level behavior. We formulate a velocity-dependent controller, based on the SMPPI algorithm, that adaptively adjusts the reference speed $v_{ref}$ according to the risk metric. This is a simple yet effective strategy and it offers intuitive engineering parameters \cite{du_velocity-dependent_2011, kahn_uncertainty-aware_2017}. The SMPPI controller was demonstrated in our prior work \cite{kim_smooth_2022} to be capable of generating optimal trajectories despite rapid changes in optimal distribution. Concretely, the reference speed decreases from a maximum speed of $v_{max} = 40$ km/h according to the following update law:
\begin{equation}
    v_{ref} = v_{max} - w_1 \log{\left(w_2 \, \Delta C_{\alpha} + w_3 \right)} ,
\end{equation}
where the parameters $w_{(\cdot)}$ are empirically tuned. Except for this risk-aware design, the same cost functions and control parameters were used across all the experiments. It is important to point out that the SMPPI controller can directly use either the FF-NN or the DPM as its forward dynamics model.

\section{Experiments}
\subsection{Experimental Setup}

The purpose of this section is to validate our idea of applying the latent proprioceptive information extracted from the physics-embedded neural network to the controller, enabling more stable driving even in challenging conditions. We designed two types of experiments.

\subsubsection{Circular track}
First, we considered a path tracking experiment on a single-lane circular track of radius $20$ m to the center of the lane. The width of the lane was $16$ m. It was designed to evaluate quantitatively whether each controller can track the reference trajectory well when continuous slipping occurs on a low-friction road. We set the center line of the track as the target trajectory and added accordingly a quadratic tracking error cost to the track cost function:
\begin{equation}
    \text{Track}(\vx) = \left(\sqrt{{p_x}^2 + {p_y}^2} - 20\right)^2 + {10000\textbf{M}\left(p_x,p_y\right)} .
\end{equation}

\subsubsection{Race track}
Second, we designed an autonomous driving experiment on a novel race track to validate the effectiveness of our risk-aware strategy. We built a new race track with a length of $790$ m in the simulator (see Fig.~\ref{fig:track}). The track consisted of two types of road surfaces, one with a friction coefficient of 1.0 and the other with a friction coefficient of 0.4.

\subsection{Experimental Results}

\begin{figure*}[t]
\centering
\includegraphics[width=0.95\textwidth]{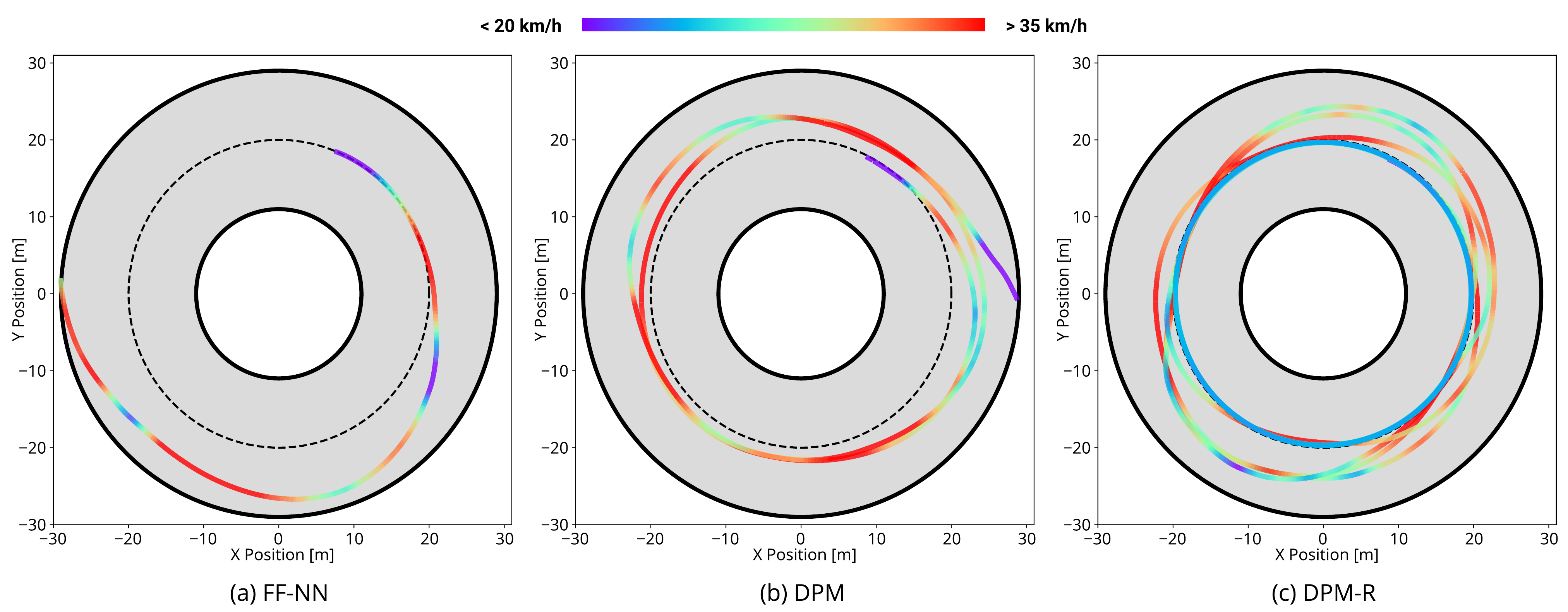}
\caption{The vehicle trajectories taken by three different control methods. The friction coefficient of the circular track was $0.45$. The black dashed lines in the center of the tracks indicate the reference trajectories.}
\label{fig:circular_track}
% \vspace*{-0.1in}
\end{figure*}

\subsubsection{Circular track}

We evaluated the control performances of the MPC using the FF-NN and the DPM. In addition, the DPM used with the risk-aware strategy (hereinafter referred to as ``DPM-R") was also evaluated. Each controller was deployed to complete 10 laps around the circular track. We ended the trial if the vehicle lost control and collided with the track boundary. 

We observed that all controllers were able to follow the reference trajectory perfectly on the roads with friction levels above $0.7$. Therefore, we conducted experiments in detail with friction levels of $\mu \in \{0.65, 0.6, 0.55, 0.5, 0.45, 0.4\}$. As an evaluation metric, the RMSE of tracking error was employed, which is the average distance the vehicle deviates from the reference trajectory. We also measured how many laps the vehicle completed without colliding with the track boundary. We referred to this as course progress rate (CPR). The experimental results for each controller are shown in Table~\ref{table:circular_error}.

 \begin{table}[ht]
\renewcommand\arraystretch{1.2}
\captionsetup{justification=centering}
\caption{}
\caption*{\textsc{Experimental results on the circular track. Average tracking errors (in [m]) and the course progress rate (CPR) with different control methods are displayed.}}
\label{table:circular_error}
\begin{center}
\begin{tabularx}{1.0\columnwidth}{c|cc|cc|cc|}
& \multicolumn{2}{c|}{FF-NN} & \multicolumn{2}{c|}{DPM} & \multicolumn{2}{c|}{DPM-R} \\ \cline{2-7}
& RMSE & CPR & RMSE & CPR & RMSE & CPR  \\ \cline{1-7}
$\mu = 0.65$ & 0.654 & 10/10 & 0.421 & 10/10 & \textbf{0.342} & 10/10 \\
$\mu = 0.6\,\,\,$ & 0.777 & 10/10 & 0.402 & 10/10 & \textbf{0.344} & 10/10 \\
$\mu = 0.55$ & 0.754 & 10/10 & 0.608 & 10/10 & \textbf{0.342} & 10/10 \\
$\mu = 0.5\,\,\,$ & 2.575 & 5/10 & 1.569 & \textbf{10/10} & \textbf{0.338} & \textbf{10/10} \\
$\mu = 0.45$ & N/A & 0/10 & 2.971 & 2/10 & \textbf{1.616} & \textbf{10/10} \\
$\mu = 0.4\,\,\,$ & N/A & 0/10 & N/A & 0/10 & \textbf{3.612} & \textbf{1/10}
\end{tabularx}
\end{center}
\vspace*{-0.15in}
\end{table}

The results show that the controller using the DPM had lower tracking errors than the one using the FF-NN at all friction levels. This is due to the fact that the physics-based model provided strong guidance at the last hidden layer of the DPM, enabling better generalization with limited training data. In particular, the DPM outperformed the FF-NN on low-friction road surfaces by a large margin. As described in Section~\ref{section:nn_result}, the DPM has a greater lateral dynamics estimation accuracy than the vanilla model, which accounts for this observation.

Even though the modeling accuracy has improved, the performance of the controller would inevitably deteriorate if the target speed remains constant despite the low-friction levels. Alternatively, the DPM-R adjusted the target speed according to the proprioceptive hints, resulting in lower tracking errors than others. It also showed a robust result in the test with a friction level of $\mu = 0.45$. This strategy is effective in cases where safety is the first priority. In the tests with $\mu = 0.4$, the friction level was too low that all controllers failed to retain their stabilities. We visualize the trajectories taken by each controller in the tests with $\mu = 0.45$ (see Fig.~\ref{fig:circular_track}). The controller using the DPM traversed a longer distance than the one using the FF-NN, but eventually collided with the track boundary. On the other hand, the DPM-R reduced the vehicle speed and successfully followed the given trajectory for 10 laps.

\begin{figure}[t]
\centering
\includegraphics[width=0.95\linewidth]{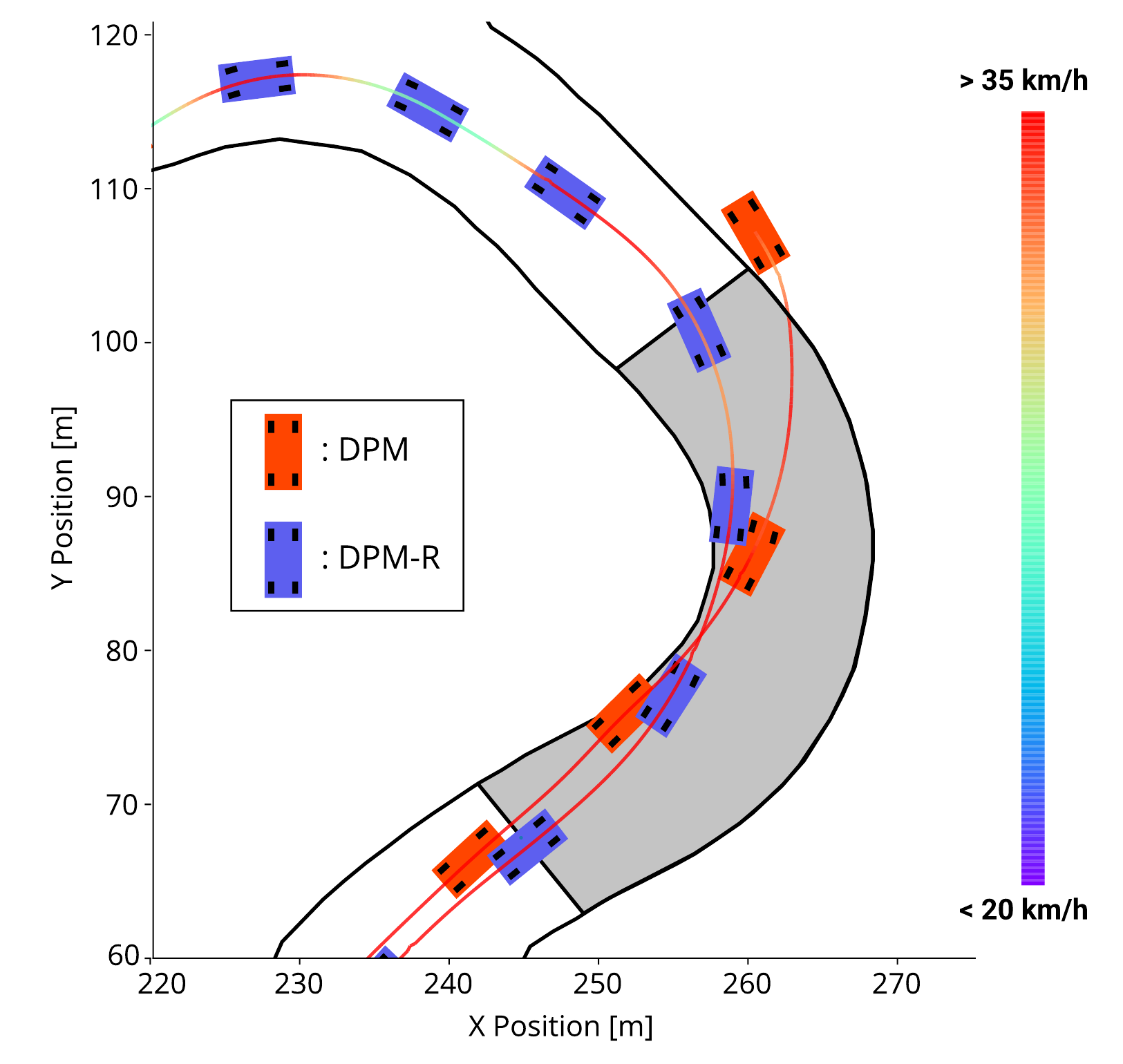}
\caption{The vehicle trajectories taken by the compared controllers. The friction coefficient of the gray region was $0.4$, and the remainder was $1.0$.}
\label{fig:corner}
\end{figure}

\subsubsection{Race track}

\begin{figure}[t]
\centering
\includegraphics[width=0.95\linewidth]{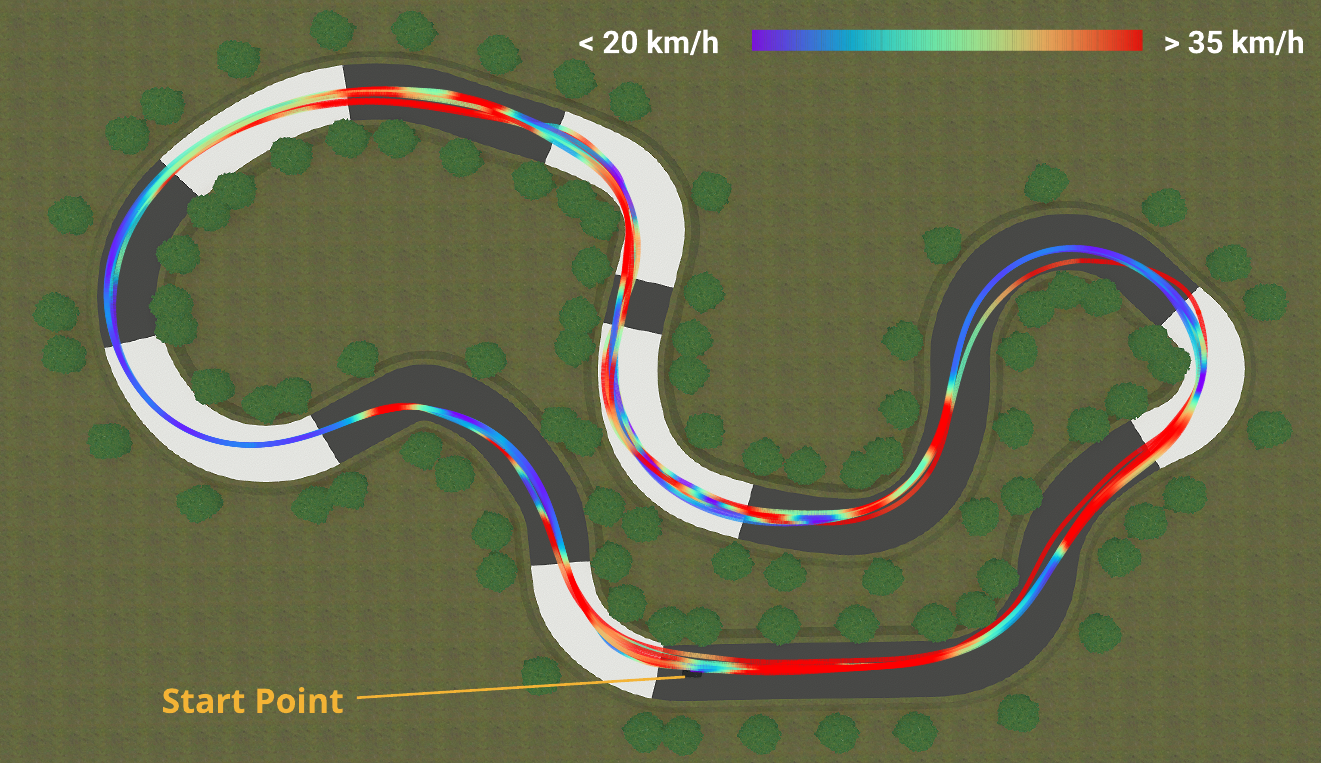}
\caption{The vehicle trajectory taken by the controller using the DPM-R. The friction coefficients of the white regions were $0.4$, and the remainder were $1.0$. }
\label{fig:track}
\end{figure}

Here, we conducted an ablation study on the DPM-R method. We visualize the vehicle trajectories taken by the controllers using the DPM and the DPM-R, respectively, while passing through the first sharp corner of the track (see Fig.~\ref{fig:corner}). The friction coefficient was suddenly changed at the corner entry ($\mu: 1.0 \to 0.4$). The sampling-based controller with the DPM attempted to respond to this corner by applying a large steering angle, but it failed because the controller's own deceleration commands were insufficient to get through this slippery corner. On the contrary, the DPM-R was informed by the latent proprioceptive information to reduce the target speed for minimizing the risky sliding motions. As a result, the proposed risk-aware controller completed 10 laps around the race track successfully (see Fig.~\ref{fig:track}). The cornering stiffness uncertainty increased as the vehicle entered the corners with low coefficients of friction, the controller then lowered the desired speed to prevent significant sliding. Subsequently, after the vehicle entered straight roads with high coefficients of friction, the controller increased the target speed again as the uncertainty variable decreased. In other words, the proposed control scheme was able to achieve both high-speed driving on high-friction surfaces and conservative driving on low-friction surfaces, by adaptively modifying the target speed based on the latent proprioceptive information.

\section{Conclusion}

We presented a physics embedded neural network vehicle model, which seamlessly combines deep learning with physics-based models, structuring the model learning problem by involving the well-known prior knowledge. The proposed model outperformed the baseline neural network model by a large margin, showing a faster learning speed, higher accuracy, and better generalization. In addition, our model is capable of accurately estimating lateral tire forces, which can be acquired from the forward propagation of the model, without the need for any additional training with ground truth data. Based on these latent features, we develop a risk-aware autonomous driving control scheme designed to achieve both high-speed driving and conservative driving depending on the environmental conditions. We conducted extensive experiments in two autonomous driving tasks with unknown friction, demonstrating that the proposed control framework shows much more robust performances than the baseline framework. Future research should focus on transferring the proposed model to a real-world vehicle using the Sim-to-Real approach.

\addtolength{\textheight}{0 cm}   % This command serves to balance the column lengths

%%%%%%%%%%%%%%%%%%%%%%%%%%%%%%%%%%%%%%%%%%%%%%%%%%%%%%%%%%%%%%%%%%%%%%%%%%%%%%%%

%%%%%%%%%%%%%%%%%%%%%%%%%%%%%%%%%%%%%%%%%%%%%%%%%%%%%%%%%%%%%%%%%%%%%%%%%%%%%%%%

%%%%%%%%%%%%%%%%%%%%%%%%%%%%%%%%%%%%%%%%%%%%%%%%%%%%%%%%%%%%%%%%%%%%%%%%%%%%%%%%

% \section*{ACKNOWLEDGMENT}
% This work was supported by Agency for Defense Development.

%%%%%%%%%%%%%%%%%%%%%%%%%%%%%%%%%%%%%%%%%%%%%%%%%%%%%%%%%%%%%%%%%%%%%%%%%%%%%%%%
\bibliographystyle{IEEEtran}
\typeout{}
\bibliography{references.bib}

\end{document}